\documentclass{article}

\usepackage[preprint]{corl_2026}

\AtBeginDocument{\nolinenumbers}
\usepackage{xcolor}
\usepackage{booktabs}
\usepackage[export]{adjustbox}
\usepackage{multirow}
\usepackage{amsmath}
\usepackage{amssymb}
\usepackage{graphicx}
\usepackage{algorithm}
\usepackage{algorithmic}
\usepackage{subcaption}
\usepackage{makecell}
\usepackage{wrapfig}
\usepackage[T1]{fontenc}
\usepackage{tikz}
\usepackage{etoolbox}

\title{UniviewVLA: A Unified Multiview Vision-Language-Action Model with World Modeling}

\author{
  \begin{minipage}{0.95\textwidth}
  \centering\bfseries
  Tao Xu$^{1,2}$,
  Runhao Zhang$^{2}$,
  Zhijian Huang$^{4}$,
  Jiayi Guan$^{4}$,
  Jiaxin Wang$^{2}$,
  Yifan Ding$^{2}$,\\
  Yong-Lu Li$^{2,3}$,
  Long Chen$^{4}$,
  Guang Chen$^{1,2,\dagger}$,
  Jinghui Lu$^{4,\dagger}$
  \end{minipage}\\[1.0em]
  \begin{minipage}{0.95\textwidth}
  \centering\normalfont\footnotesize
  $^{1}$Tongji University,
  $^{2}$Shanghai Innovation Institute,
  $^{3}$Shanghai Jiao Tong University,
  $^{4}$Xiaomi EV
  \end{minipage}\\[-2.8em]
}

\makeatletter
\renewcommand{\@notice}{}
\patchcmd{\@maketitle}
  {\vskip 0.3in \@minus 0.1in}
  {\vskip 0.12in \@minus 0.05in}
  {}{}
\makeatother

\begin{document}

\maketitle

\begin{tikzpicture}[remember picture,overlay]
  \node[anchor=south west, inner sep=0]
    at ([xshift=\dimexpr1in+\oddsidemargin\relax,yshift=0.55in]current page.south west) {%
\parbox{\textwidth}{\footnotesize\raggedright
$^{\dagger}$Corresponding authors.}%
  };
\end{tikzpicture}

\begin{abstract}
Occluded tasks remain a bottleneck in robot manipulation. Existing solutions either deploy additional physical cameras requiring training-inference camera parity, or rely on explicit 3D reconstruction with high computational cost. Moreover, both approaches rely on standard agent-view and wrist-view observations, while failing to capture occlusion information and future scene evolution. To this end, we propose \textbf{UniviewVLA}, a \textbf{uni}fied multi\textbf{view} \textbf{V}ision-\textbf{L}anguage-\textbf{A}ction model with world modeling, which infers multiview scene evolution for action prediction from only standard two-camera observations. We demonstrate that by leveraging generated multiview future views from the world model, UniviewVLA reveals occluded cues and models future scene evolution, improving action prediction and removing the need for extra hardware or explicit reconstruction. Besides, to accelerate inference while preserving prediction accuracy, UniviewVLA develops Motion-Informative Token Compression, which compresses each generated view from 625 to 16 tokens and reduces per-view latency from 6--7s to 0.2--0.3s. UniviewVLA also proposes training-free Action-Entropy View Selection, which dynamically identifies the most action-informative view at different inference stages. Extensive experiments show that UniviewVLA achieves 95.8\% on LIBERO and 4.60 on CALVIN ABCD$\to$D, both standard occlusion-free benchmarks. On customized occlusion-focused tasks, it improves success rate from 40.0\% to 73.3\%, and average real-robot success rate by 33.4 points, demonstrating stronger occlusion-focused performance without sacrificing standard occlusion-free benchmarks. Project website: \href{https://sii-quantum.github.io/MultiviewVLA.github.io/}{This URL}.

\end{abstract}

% \keywords{Multiview Robot Learning, Vision-Language-Action Models,
%           Generative World Models, Dynamic View Selection}

% %===========================================================================
\section{Introduction}

Vision-Language-Action (VLA) models have made substantial progress in robot manipulation by mapping visual observations and language instructions to robot actions~\citep{zitkovich2023rt, kim2025openvla, black2024pi_0, wang2025unified, fan2025long}. However, their performance is fundamentally constrained by the visual information available at deployment time. In manipulation, action-critical cues, including gripper-object relations, object poses, and contact regions, are highly view-dependent. When these cues are occluded in the deployed camera views, the model may fail even if the underlying task knowledge has been learned.

A natural solution is to expand the visual input by deploying additional physical cameras, and prior multiview methods have shown that richer visual coverage improves manipulation performance~\citep{cheang2024gr, li2026multi, xie2026multi}. However, these approaches require strict camera parity between training and inference, \textit{i.e.,} the number, placement, and calibration of cameras must be consistently replicated across deployment sites, which is brittle and difficult to scale.  Another line of work resorts to explicit 3D reconstruction to infer scene geometry from partial observations~\citep{wen2025dexvla, goyal2024rvt, goyal2023rvt, shridhar2023perceiver}, but it is limited to current scene states and incurs substantial computational cost.  This motivates a central question: \textbf{Can a world model conditioned on only standard observations infer multiview information and future scene evolution for downstream action prediction, and thereby improve occlusion-focused manipulation without high-cost sensing setups?}

To answer this question, we propose \textbf{UniviewVLA}, a \textbf{uni}fied multi\textbf{view} \textbf{VLA} model with world modeling capabilities, that infers multiview scene evolution for action prediction. The model learns to predict multiview future scene evolution from only standard two-camera observations. By conditioning action prediction on the generated auxiliary future-view representations, UniviewVLA injects multiview information and future scene evolution knowledge into robot manipulation, and produces occluded action-critical evidence without requiring costly additional camera deployment or explicit 3D reconstruction~\citep{hafner2019dream,wu2023daydreamer,yang2026chain,cai2026beyond,tian2024view,wang2026efficient}. We first demonstrate that introducing additional views, whether derived from ground-truth images quantized using visual quantization (VQ) techniques~\cite{wang2025unified} or generated by a world model, consistently improves prediction accuracy, as shown in Section~\ref{sec:pre_exp}. Further, we discovered that such formulation introduces two practical challenges: (1) redundant auxiliary-view tokens increase latency and may harm action prediction, and (2) the most informative viewpoint varies across different inference stages. UniviewVLA addresses token redundancy with \emph{Motion-Informative Token Compression} (Section~\ref{sec:compact_action}), which compresses each generated view from 625 to 16 tokens and reduces per-view latency from 6--7s to 0.2--0.3s. To enable dynamic view selection, UniviewVLA also proposes \emph{Action-Entropy View Selection} (Section~\ref{sec:entropy}), which identifies the most action-informative view for action prediction. We evaluate UniviewVLA on standard occlusion-free benchmarks, where it achieves 95.8\% success on LIBERO and 4.60 on CALVIN ABCD$\to$D. To further evaluate its occlusion performance, we construct six occlusion-focused simulation tasks with customized occlusion scenarios and collect multiview demonstrations through 3D-SpaceMouse teleoperation. UniviewVLA achieves an average improvement from 40.0\% to 73.3\% success on these tasks. In two real-world occlusion tasks, it yields a 33.4 points gain in average success rate.

Our contributions are summarized as follows:

\begin{itemize}

\item We propose \textbf{UniviewVLA}, a unified multiview VLA framework with world modeling that infers multiview future scene evolution before predicting action, without requiring additional physical cameras or explicit 3D reconstruction. Experimental results show that UniviewVLA achieves an average improvement of 33.3 percentage points on six occlusion-focused simulation tasks and 33.4 percentage points gain on two real-world occlusion-focused tasks.

\item We develop Motion-Informative Token Compression and Action-Entropy View Selection to reduce redundant auxiliary-view tokens and enable dynamic multiview selection across inference stages without additional training, significantly decreasing the inference latency.

\item We introduce six customized occlusion-focused manipulation tasks and an open-source pipeline for multiview data collection, processing, and evaluation, enabling assessment of occlusion-focused performance.

\end{itemize}

\section{Related Work}
\label{sec:related}

\paragraph{Vision-Language-Action Models.}
Modern Vision-Language-Action (VLA) models incorporate language conditioning and vision-language pretraining~\citep{reuss2024multimodalmdt,huang2025robotron,song2026fastvla,zhaolearning,chi2025diffusion,brohan2023rt,huang2024making}, leading to generalist policies that map language-conditioned visual observations directly to robot actions~\citep{ahn2022can,driess2023palm,zitkovich2023rt,li2024vision,mees2024octo,kim2025openvla,bousmalis2023robocat,o2024open}. Recent work further improves action prediction performance through flow matching, action tokenization, diffusion experts, video-based policy learning, and world modeling~\citep{black2024pi_0,pertsch2025fast,wen2025dexvla,cheang2024gr,wang2025unified,fan2025long}. Yet most models still perceive the scene through a fixed set of deployed cameras. When action-critical cues are occluded beyond these fixed views, policy-side architectural improvements alone cannot overcome the resulting observability bottleneck.

\paragraph{Multiview Robot Learning.}
Multiview perception has been widely adopted in robot manipulation to provide complementary visual evidence for robot learning~\citep{james2020rlbench,shridhar2023perceiver,goyal2023rvt,goyal2024rvt,xie2026multi}. These methods exploit multiview observations through voxelized representations, virtual re-rendering, multiview transformers, view-scaled demonstrations, or generated views and demonstrations~\citep{zeng2021transporter,shridhar2022cliport,huang2023voxposer,ke20253d,cai2026beyond,yang2025novel,li2026multi}. However, their gains are often tied to hardware constraints: policies trained with additional views typically require similar camera poses, calibration, and synchronization at deployment. 

%===========================================================================

\section{Preliminary Experiments}\label{sec:pre_exp}

\vspace{-1.5mm}

\begin{figure}[h]
  \centering
  \includegraphics[width=\linewidth]{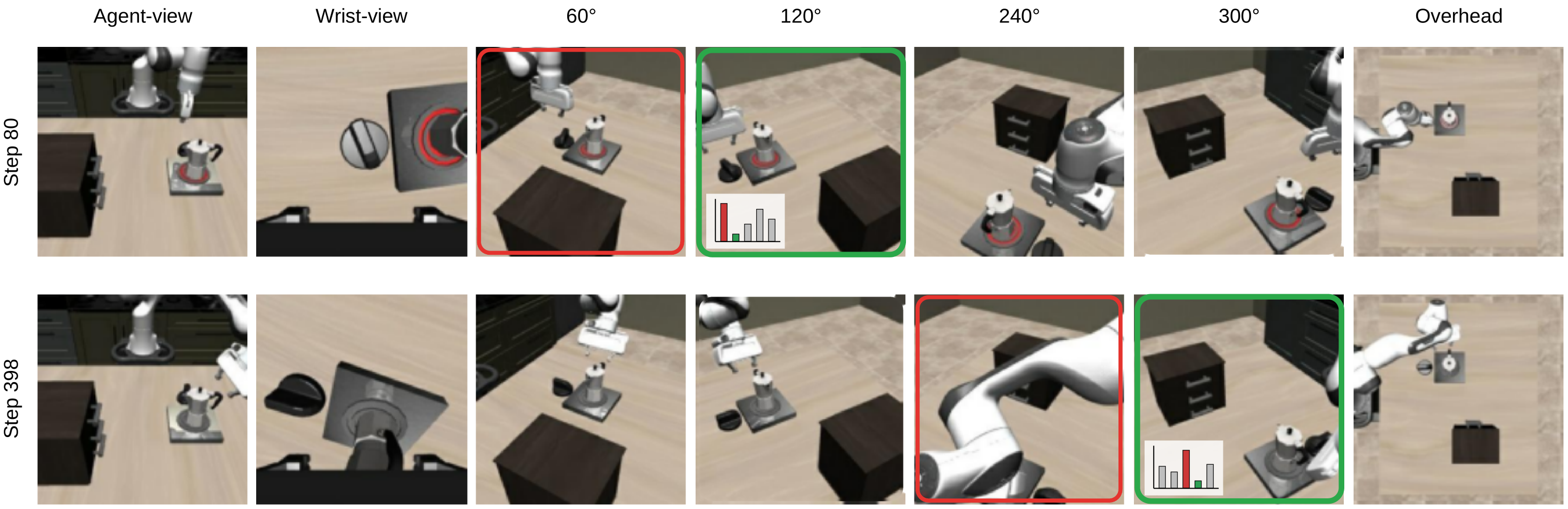}
\caption{
\textbf{Multiview observations under occlusion.}
Green boxes mark the selected best views with the lowest action entropy, while red boxes mark higher-entropy views. Bar charts show action entropy, with lower entropy indicating a more action-informative view.
}
  \label{fig:mask_example}
\end{figure}

At deployment, VLA models usually use only two standard physical observations, the agent-view $\mathbf{x}_a$ and wrist-view $\mathbf{x}_w$, which may miss critical information under occlusions. As shown in Fig.~\ref{fig:mask_example} and Table~\ref{tab:occlusion_view_selection}, the agent-view cannot observe the occluded switch in Task~6, whereas the $120^\circ$ third-camera view exposes it, improving success from $4\%$ to $16\%$ and highlighting the value of multiview information for occlusion-focused manipulation. Nevertheless, physical multiview policies require matched camera configurations between training and inference, and only observe the current scene without predicting future scene evolution. UniviewVLA addresses these constraints by generating auxiliary future views from the two standard observations.

\vspace{-1.5mm}
\vspace{-1.5mm}

\begin{wrapfigure}[12]{r}{0.34\linewidth}
\vspace{-6pt}
\centering
\includegraphics[width=0.88\linewidth]{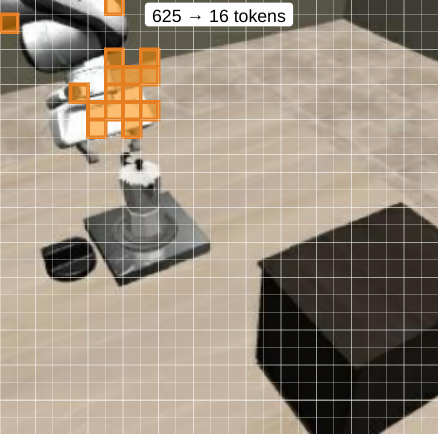}
\vspace{-4pt}
\captionsetup{font=small}
\caption{\textbf{Motion-informative token compression.}}
\label{fig:compact_tokens}
\vspace{-10pt}
\end{wrapfigure}

However, as shown in Figure~\ref{fig:compact_tokens}, each auxiliary-view generation incurs excessive visual tokens from dense future-view prediction, leading to high inference latency and redundant information that may impair action prediction. Therefore, compact auxiliary views that preserve motion-informative evidence are important for downstream action prediction. 

Moreover, the best viewpoint may change across different inference stages. As shown in Fig.~\ref{fig:mask_example}, the bar charts report action entropy across auxiliary views, where lower entropy indicates a more action-informative view. For example, the $120^\circ$ view reveals the occluded switch more clearly than the $300^\circ$ view at step~80, whereas the $300^\circ$ view better exposes the mug handle at step~398. The $240^\circ$ view further shows this temporal shift: it is informative early but becomes self-occluded later. These observations motivate dynamic action-informative view selection.

%===========================================================================

\section{Method}\label{sec:method}

\begin{figure*}[t]
  \centering
  \includegraphics[width=\textwidth]{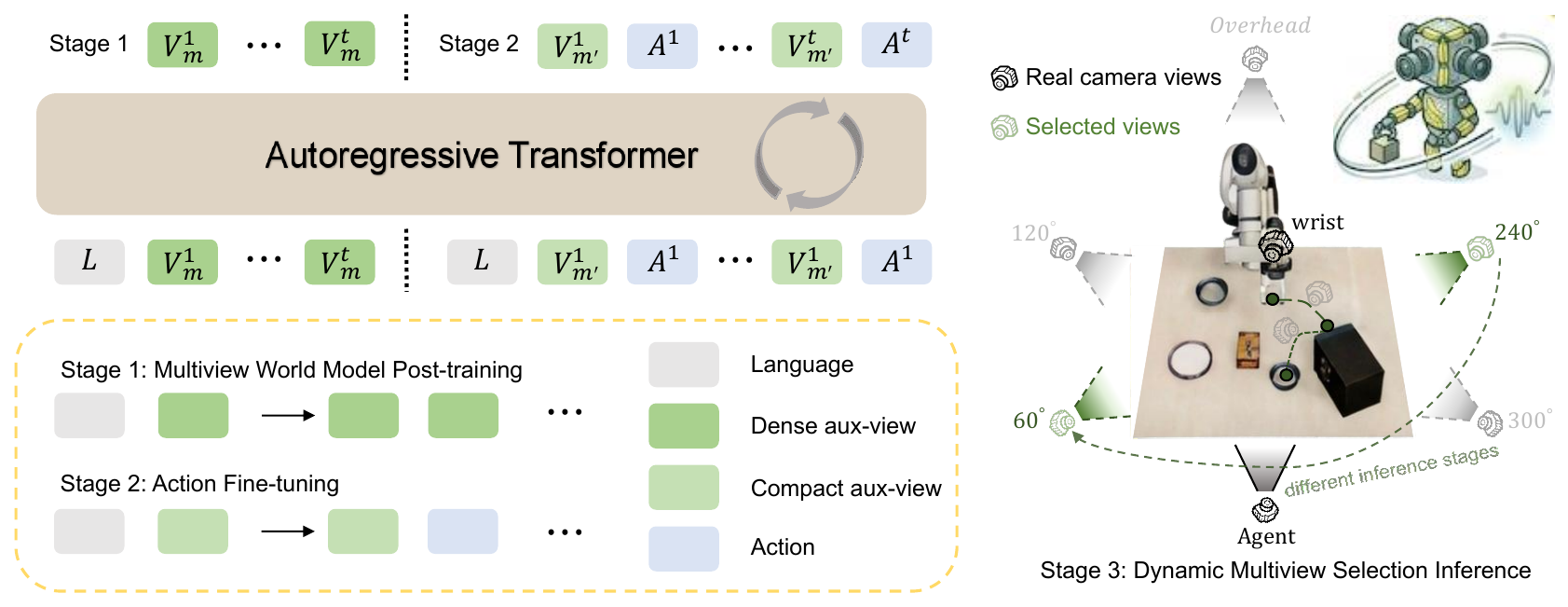}
\caption{%
\textbf{UniviewVLA pipeline.}
UniviewVLA models language instructions, multiview observations, and actions with discrete tokens that can be autoregressively predicted by a unified Transformer model~\cite{wang2025unified}, using two training stages and dynamic inference.
\textbf{(1)~Multiview world model post-training.}
UniviewVLA takes language instructions, standard agent-view, and wrist-view inputs, and autoregressively generates future multiview images that incorporate multiview and world evolution information.
\textbf{(2)~Action fine-tuning.}
UniviewVLA first predicts compact motion-informative tokens to avoid the high latency of full auxiliary-view tokens, and then predicts FAST action tokens.
\textbf{(3)~Dynamic inference.}
During inference, transparent cameras denote generated auxiliary views, and the green camera denotes the selected auxiliary view. UniviewVLA periodically selects the best auxiliary view for action prediction across different inference stages instead of using a fixed viewpoint.}

  \label{fig:overview}
\end{figure*}

\subsection{Overview}
\label{sec:overview}

As illustrated in Figure~\ref{fig:overview}, UniviewVLA follows the discrete-token autoregressive modeling paradigm of UniVLA~\citep{wang2025unified}, representing language, visual observations, auxiliary views, and actions as token sequences modeled by an autoregressive transformer $\mathcal{M}_\theta$. Given the standard agent-view $\mathbf{x}_a$, wrist-view $\mathbf{x}_w$, and instruction $L$, we denote by $o_{t-1:t}=\{\mathbf{x}_{a,t-1:t},\mathbf{x}_{w,t-1:t}\}$ the two-frame standard observation history, and by $\mathcal{V}_{\mathrm{sel}}$ the candidate auxiliary-view set. UniviewVLA is trained in two stages:
\vspace{-1.5mm}
\begin{equation}
\begin{aligned}
& \text{(Stage 1) World model post-training:} \quad
\hat{o}_{t+1}^v \sim p_\theta(\cdot \mid o_{t-1:t},\, L),
\quad v \in \mathcal{V}_{\mathrm{sel}}, \\
& \text{(Stage 2) Action fine-tuning:} \quad
\hat{o}_{t+1}^{\prime v},\, a_t \sim \pi_\theta(\cdot \mid o_{t-1:t},\, L),
\quad v \in \mathcal{V}_{\mathrm{sel}}.
\end{aligned}
\label{eq:two_stage_training}
\vspace{-1.5mm}
\end{equation}
Here, $\hat{o}_{t+1}^v$ denotes the generated future auxiliary-view representation, $\hat{o}_{t+1}^{\prime v}$ denotes its compact motion-informative form, and $a_t$ denotes the action token predicted from this representation.

\subsection{Generative Multiview World Model Post-training}
\label{sec:worldmodel} 
In Stage~1, the world model learns to generate future auxiliary views from the two-frame standard physical observations, providing both multiview information and predictive scene evolution knowledge for downstream action prediction. Specifically, given $o_{t-1:t}=\{\mathbf{x}_{a,t-1:t},\mathbf{x}_{w,t-1:t}\}$ and instruction $L$, it predicts the future VQ-token sequence $\mathbf{c}_{t+1}^v=(c_{t+1,1}^v,\ldots,c_{t+1,N_{\mathrm{tok}}}^v)$ for each auxiliary view $v\in\mathcal{V}_{\mathrm{sel}}$, where $N_{\mathrm{tok}}$ denotes the number of VQ tokens per view. The training objective is autoregressive visual prediction:
\begin{equation}
\mathcal{L}_{\mathrm{world}}
= - \sum_{v\in\mathcal{V}_{\mathrm{sel}}}
\sum_{i=1}^{N_{\mathrm{tok}}}
\log p_\theta\left(
c_{t+1,i}^v
\mid
c_{t+1,<i}^v,\,
o_{t-1:t},\, L
\right),
\label{eq:worldmodel_loss}
\end{equation}
where $c_{t+1,<i}^v$ denotes the previously generated tokens of the future auxiliary view from viewpoint $v$. By supervising future auxiliary-view prediction, this objective equips the model with future multiview prediction ability and injects scene evolution knowledge into the generated view tokens.

\subsection{Joint Motion-Informative Token Compression and Action Fine-tuning}
\label{sec:compact_action}

As shown in Figure~\ref{fig:compact_tokens}, each full auxiliary view contains $N_{\mathrm{tok}}=25\times25=625$ VQ tokens, and generating five views requires $3125$ tokens per step. To reduce this cost, we retain motion-relevant tokens from each predicted view. Given the VQ-token sequences $\mathbf{c}_{t}^v$ and $\mathbf{c}_{t+1}^v$ of auxiliary view $v$ at consecutive timesteps $t$ and $t+1$, we compute a cosine-distance score for each spatial token $j$:
\begin{equation}
\delta_{t,j}^v
= 1 - \cos\left(E(c_{t,j}^v),\ E(c_{t+1,j}^v)\right),
\quad j = 1, \ldots, N_{\mathrm{tok}} ,
\end{equation}

where $E(\cdot)$ denotes the VQ embedding table. A higher $\delta_{t,j}^v$ indicates stronger visual change and therefore higher motion relevance. We retain the top $K=16$ tokens ranked by $\delta_{t,j}^v$, yielding the compact motion-informative representation $\hat{o}_{t+1}^{\prime v}$ for each view. The model is trained to predict these compact tokens and directly generates them at inference. This reduces each auxiliary view from 625 to 16 tokens and the total auxiliary-view token budget from 3125 to 80.

Given the compact representation, action fine-tuning concatenates the $K=16$ tokens of each candidate view with the action-token sequence $\mathbf{a}_t=(a_{t,1},\ldots,a_{t,N_a})$:
\[
\mathbf{s}_t^v =
\left(
\hat{o}_{t+1,1}^{\prime v}, \ldots, \hat{o}_{t+1,K}^{\prime v},
a_{t,1}, \ldots, a_{t,N_a}
\right),
\]
where $N_a$ is the number of action tokens. The training objective applies autoregressive cross-entropy supervision over the concatenated sequence:
\begin{equation}
\mathcal{L}_{\mathrm{act}}
= - \sum_{v\in\mathcal{V}_{\mathrm{sel}}}
\sum_{m=1}^{K + N_a}
\log p_\theta\!\left(
s_{t,m}^v
\mid
s_{t,<m}^v,\,
o_{t-1:t},\, L
\right).
\label{eq:action_loss}
\end{equation}
Each candidate view is trained as a view-prefixed sequence, enabling action prediction to condition on compact multiview evidence and the scene evolution knowledge introduced by the world model.

\subsection{Dynamic Test-Time View Selection via Action Entropy}
\label{sec:entropy}

As shown in Fig.~\ref{fig:overview} and motivated by Section~\ref{sec:pre_exp}, UniviewVLA dynamically selects the most action-informative auxiliary view during inference instead of relying on a fixed generated viewpoint. For each view-selection round, it specifies a candidate view $v\in\mathcal{V}_{\mathrm{sel}}$ to the model and generates the corresponding compact representation $\hat{o}_{t+1}^{\prime v}$ from $(o_{t-1:t},L)$. We then compute the mean action-token entropy for this candidate view:
\begin{equation}
H_v =
\frac{1}{N_a}\sum_{n=1}^{N_a}
\mathcal{H}\left[
p_\theta(a_{t,n} \mid a_{t,<n},\hat{o}_{t+1}^{\prime v},
o_{t-1:t},L)\right].
\label{eq:h_view}
\end{equation}

where $a_{t,n}$ denotes the $n$-th token in the action-token sequence $\mathbf{a}_t$, and $\mathcal{H}[p]=-\sum_a p(a)\log p(a)$ denotes Shannon entropy over the action-token distribution~\citep{shannon1948mathematical}. UniviewVLA identifies the most action-informative view as the one with the lowest action entropy:

\begin{equation}
v^*=\arg\min_{v\in\mathcal{V}_{\mathrm{sel}}} H_v,\qquad
\hat{o}_{t+1}^{\prime *}=\hat{o}_{t+1}^{\prime v^*}.
\label{eq:view_selection}
\end{equation}
The final action-token sequence is predicted conditioned on $(o_{t-1:t},L,\hat{o}_{t+1}^{\prime *})$, leveraging compact multiview evidence and future scene evolution knowledge. To adapt to stage-dependent viewpoint changes, UniviewVLA dynamically updates the selected view every 30 timesteps, requiring no view-selection labels or additional training objective.

\section{Experiments}
\subsection{Experimental Setup}
\label{sec:exp_protocol}

We first evaluate UniviewVLA on standard occlusion-free simulation benchmarks, including LIBERO~\citep{liu2023libero} and CALVIN ABCD$\to$D~\citep{mees2022calvin}. To further evaluate occlusion-focused manipulation, we introduce six LIBERO-style occlusion tasks and two real-robot occlusion tasks on an ALOHA platform. For the simulation occlusion tasks, we collect demonstrations through 3D-SpaceMouse teleoperation. In both simulation and real-robot occlusion tasks, action-critical cues are occluded from the agent-view and wrist-view cameras, allowing us to evaluate whether generated auxiliary views improve occlusion-aware robot manipulation. At inference time, UniviewVLA uses only the agent-view and wrist-view as physical camera inputs. Implementation details are provided in Appendix~\ref{sec:appendix_implementation}, and task details are provided in Appendix~\ref{Tasks Details}.

\begin{table}[h]
  \centering
  \caption{%
    \textbf{Long-horizon manipulation performance on CALVIN ABCD$\to$D.}
  }
  \label{tab:calvin}
  \small
  \setlength{\tabcolsep}{5pt}
  \begin{tabular*}{0.94\linewidth}{@{\extracolsep{\fill}}lcccccc}
    \toprule
    Method & 1 & 2 & 3 & 4 & 5 & Avg. \\
    \midrule
    RT-1~\cite{brohan2023rt}           & 0.844 & 0.617 & 0.438 & 0.323 & 0.227 & 2.45 \\
    Robo-Flamingo~\cite{li2024vision}  & 0.964 & 0.896 & 0.824 & 0.740 & 0.660 & 4.09 \\
    GR-1~\cite{wu2024unleashing}       & 0.949 & 0.896 & 0.844 & 0.789 & 0.731 & 4.21 \\
    MDT~\cite{reuss2024multimodalmdt}  & \textbf{0.986} & \textbf{0.958} & 0.916 & 0.862 & 0.801 & 4.52 \\
    RoboVLMs~\cite{liu2025towards}     & 0.967 & 0.930 & 0.899 & 0.865 & 0.826 & 4.49 \\
    Fast-dVLA~\cite{song2026fastvla}   & 0.984 & 0.952 & 0.922 & 0.870 & 0.812 & 4.54 \\
    MINT~\cite{mint_huang2026mimic}    & 0.974 & 0.942 & 0.917 & 0.882 & \textbf{0.861} & 4.57 \\
    \textbf{UniviewVLA}                         & 0.983 & \textbf{0.958} & \textbf{0.928}
                                        & \textbf{0.893} & 0.838 & \textbf{4.60} \\
    \bottomrule
  \end{tabular*}
\end{table}

\begin{table}[h]
  \centering
\caption{%
  \textbf{Manipulation success rates on the LIBERO benchmark.}
}
  \label{tab:libero}
  \small
  \setlength{\tabcolsep}{5pt}
  \begin{tabular*}{0.94\linewidth}{@{\extracolsep{\fill}}lccccc}
    \toprule
    Method & Long & Goal & Spatial & Object & Avg. \\
    \midrule
    OpenVLA~\cite{kim2025openvla}       & 53.7 & 79.2 & 84.9 & 88.4 & 76.6 \\
    $\pi_0$-FAST~\cite{pertsch2025fast} & 60.2 & 88.6 & 96.4 & 96.8 & 85.5 \\
    CoT-VLA~\cite{zhao2025cotvla}       & 69.0 & 87.6 & 87.5 & 91.6 & 83.9 \\
    VLA-0~\cite{goyal2025vla0}          & 87.6 & \textbf{96.2} & \textbf{97.0} & 97.8 & 94.7 \\
    GR00T N1~\cite{bjorck2025gr00t}     & 90.6 & 93.0 & 94.4 & 97.6 & 93.9 \\
    UniVLA~\cite{wang2025unified}       & 91.4 & 93.2 & 96.0 & \textbf{99.2} & 95.0 \\
    OpenVLA-OFT~\cite{kim2025fine-openvla-oft} & 90.7 & 96.2 & 96.2 & 98.3 & 95.4 \\
    \textbf{UniviewVLA}                          & \textbf{94.2} & 93.4 & 96.4 & \textbf{99.2} & \textbf{95.8} \\
    \bottomrule
  \end{tabular*}
\end{table}

\subsection{Comparison with State-of-the-Art Methods}
\label{sec:benchmark_performance}

\textbf{CALVIN.} Table~\ref{tab:calvin} reports CALVIN ABCD$\to$D results. UniviewVLA achieves the highest average chain length among the compared methods. This demonstrates its effectiveness in long-horizon, multi-task manipulation, where the world model's generated future views provide additional visual evidence and predictive dynamics knowledge across extended action sequences.

\textbf{LIBERO.} Table~\ref{tab:libero} compares UniviewVLA with state-of-the-art methods on LIBERO. UniviewVLA achieves the best average success rate of 95.8\% over the four standard suites, evaluated over 500 episodes per suite. It outperforms GR00T N1 and $\pi_0$-FAST by 1.9 and 10.3 percentage points, respectively, showing that generated multiview evidence from the world model also benefits standard occlusion-free manipulation beyond the targeted occlusion benchmarks.

\begin{figure}[h]
  \centering
  \includegraphics[width=\linewidth]{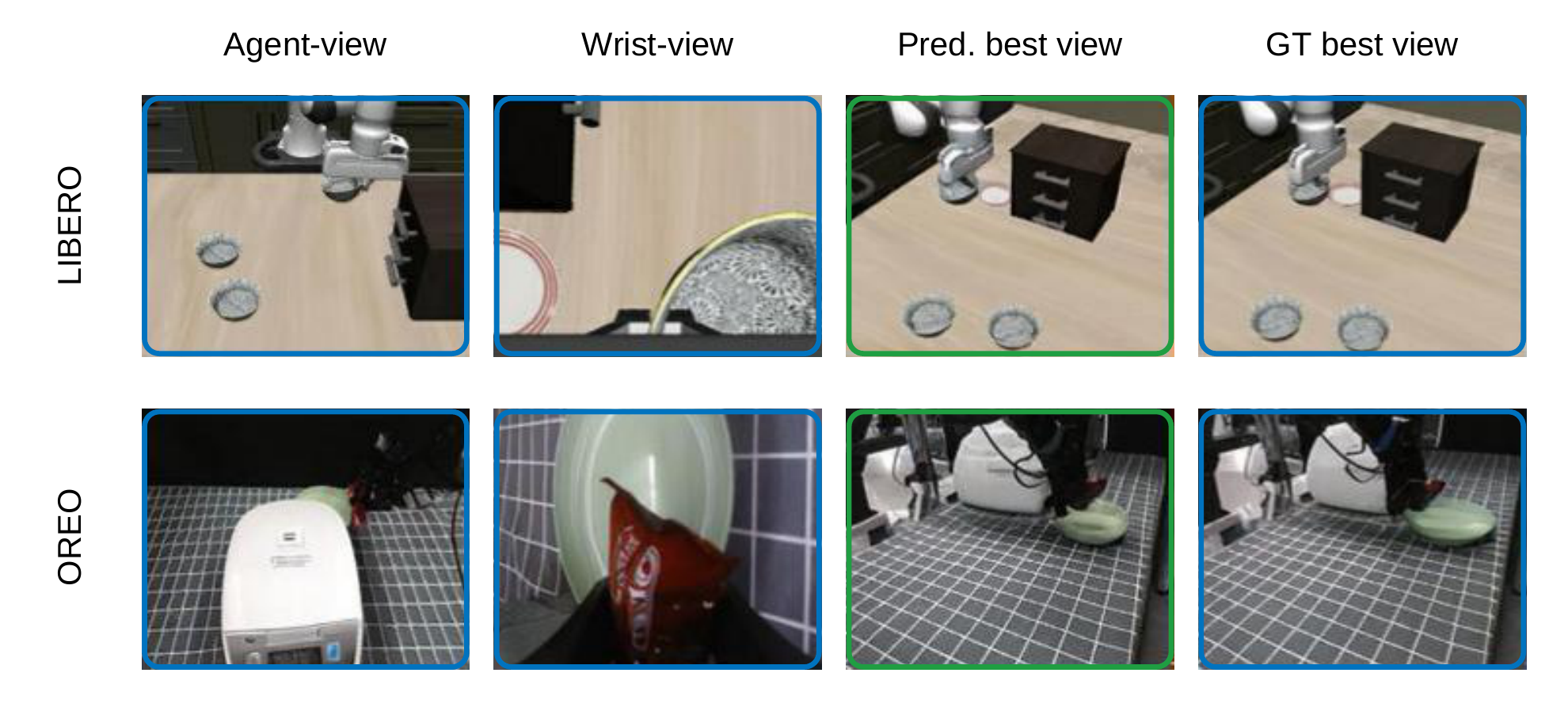}
\caption{
\textbf{Full future auxiliary-view token generation.}
The first two blue boxed columns denote the standard physical camera views (agent-view and wrist-view). Green boxes denote the generated future auxiliary view selected from these inputs. The last blue boxed column shows the ground-truth physical camera observation from the same auxiliary viewpoint for comparison.
}
  \label{fig:wm_prediction}
\end{figure}

\begin{table}[h]
  \centering
  \caption{
  \textbf{Camera-configuration evaluation on standard occlusion-free simulation benchmarks.}
  Success rates are reported on LIBERO and average chain length on CALVIN ABCD$\to$D.
    \textbf{A}: agent-view, \textbf{W}: wrist-view, \textbf{S}: physical side-view.
  }
  \label{tab:libero_calvin}
  \small
  \setlength{\tabcolsep}{5pt}
  \begin{tabular*}{\linewidth}{@{\extracolsep{\fill}}llcccccc}
    \toprule
    Method & Cameras
    & \makecell{LIBERO\\Long}
    & \makecell{LIBERO\\Goal}
    & \makecell{LIBERO\\Spatial}
    & \makecell{LIBERO\\Object}
    & \makecell{LIBERO\\Avg.}
    & \makecell{CALVIN\\Avg.} \\
    \midrule
    Two-Camera Policy    & A+W   & 90.2 & 93.4 & 92.8 & 95.6 & 93.0 & 4.42 \\
    Three-Camera Policy  & A+W+S & \textbf{94.2}
    & \textbf{95.4} & 93.6 & 97.4 & 95.2 & 4.54 \\
    \textbf{UniviewVLA}           & A+W   & \textbf{94.2} & 93.4 & \textbf{96.4}
                          & \textbf{99.2} & \textbf{95.8} & \textbf{4.60} \\
    \bottomrule
  \end{tabular*}
\end{table}

\subsection{Analysis of Generated Views without Camera Coupling}

\label{sec:4.3}

This section evaluates whether world model generated multiview evidence can replace additional deployment-time cameras. Figure~\ref{fig:wm_prediction} visualizes full future auxiliary-view generation from standard agent-view and wrist-view inputs, where the world model predicts future dynamics based on standard observations. Full auxiliary-view generation requires $25\times25=625$ tokens per viewpoint and incurs 6--7 seconds of latency, while UniviewVLA reduces each generated view to $4\times4=16$ tokens through Motion-Informative Token Compression, lowering per-view latency to 0.2--0.3 seconds. Table~\ref{tab:libero_calvin} compares three view-input configurations. During both training and inference, the \emph{Two-Camera Policy} uses only the agent-view and wrist-view, while the \emph{Three-Camera Policy} additionally uses a physical camera at a manually selected viewpoint. Both camera baselines follow the two-stage training design with future-view prediction and action fine-tuning, controlling for world-evolution knowledge and focusing on the effect of compact generated auxiliary-view tokens.

As shown in Table~\ref{tab:libero_calvin}, additional visual coverage improves standard benchmark performance. The physical third camera raises the LIBERO average from 93.0\% to 95.2\%, while UniviewVLA reaches 95.8\% with only two physical cameras. On CALVIN, UniviewVLA improves over both the two-camera policy (4.42) and the three-camera policy (4.54). These results suggest that compact world model generated workspace views provide the performance gains of additional viewpoint coverage without the deployment cost of an extra camera, while entropy-based dynamic selection further avoids the limitation of relying on a single fixed side viewpoint.

\subsection{Occlusion-Focused Evaluation and Dynamic View Selection}
\label{sec:occlusion_eval}

\begin{table}[h]
  \centering
\caption{
  \textbf{Occlusion-focused evaluation with camera and view-selection variants.}
  Success rates (\%) are reported on six customized LIBERO-style occlusion tasks.
}
  \label{tab:occlusion_view_selection}
  \small
  \setlength{\tabcolsep}{5pt}
  \begin{tabular*}{\linewidth}{@{\extracolsep{\fill}}lccccccc}
    \toprule
    Method
      & \makecell{Task 1\\}
      & \makecell{Task 2\\}
      & \makecell{Task 3\\}
      & \makecell{Task 4\\}
      & \makecell{Task 5\\}
      & \makecell{Task 6\\}
      & Avg. \\
    \midrule
    Two-Camera Policy       & 24 & 68 & 34 & 26 & 84 & 4  & 40.0 \\
    Three-Camera Policy     & 52 & 74 & 68 & \textbf{88} & \textbf{100} & 16 & 66.3 \\
    \textbf{UniviewVLA} (Manual)     & 48 & 78 & 66 & 86 & 96 & 28 & 67.0 \\
    \textbf{UniviewVLA} (Entropy)    & \textbf{58} & \textbf{88} & \textbf{72} & \textbf{88} & 98 & \textbf{36} & \textbf{73.3} \\
    \bottomrule
  \end{tabular*}
\end{table}

The occlusion tasks further evaluate whether world model generated views help when critical state information is hidden from the deployed agent-view and wrist-view cameras. As shown in Table~\ref{tab:occlusion_view_selection}, the two-camera policy achieves only 40.0\% average success, and achieves only 4\% on Task 6 where the switch is heavily occluded. This confirms that the default deployed views are insufficient when key task information is visually inaccessible. Adding a physical third camera improves the average to 66.3\%, while UniviewVLA reaches 67.0\% with a manually selected best fixed view, matching the fixed viewpoint of the Three-Camera Policy, and 73.3\% with entropy-based dynamic selection.

These results demonstrate that compact world model generated views that predict future dynamics based on standard observations can outperform adding a physical camera by efficiently predicting additional viewpoint evidence without the cost of extra camera deployment. Entropy-based selection further adapts this evidence to stage-dependent occlusions. 

\subsection{Real-Robot Deployment}
\label{sec:real_robot}

\begin{figure}[h]
  \centering
  \includegraphics[width=\linewidth]{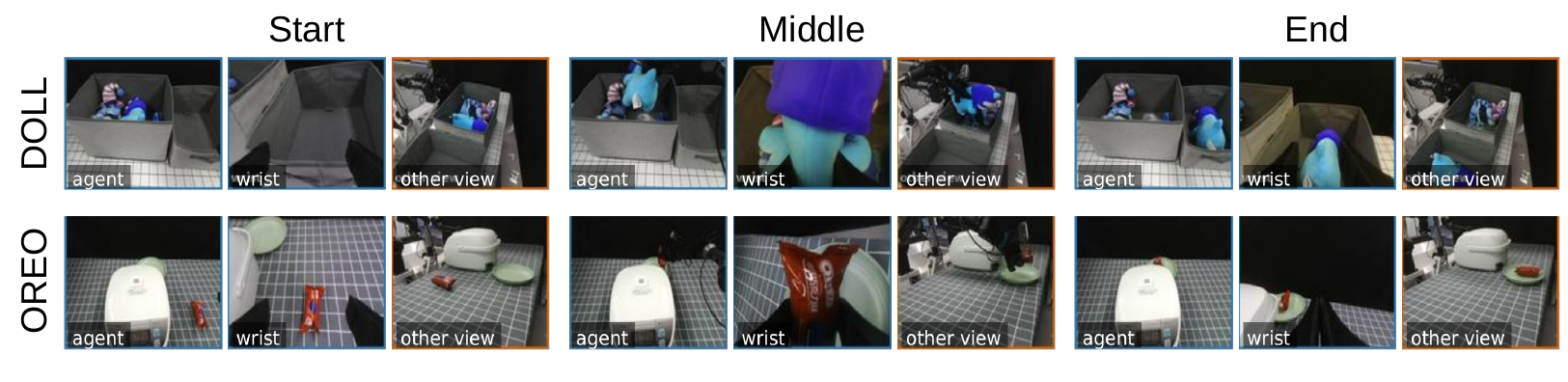}
  \caption{%
    \textbf{Real-robot occlusion tasks.}
    The target plate or manipulated object is partially hidden from the default
    agent-view, requiring additional spatial evidence for reliable execution.
  }
  \label{fig:real_robot_tasks}
\end{figure}

In addition, UniviewVLA is deployed on an ALOHA platform for real-world occlusion evaluation. The setup includes two right-arm-only manipulation tasks, each evaluated over 15 trials. To enable a practical multiview comparison with minimal hardware modification, the unused left-wrist camera is detached and extended to provide an additional physical viewpoint, allowing comparison among two-camera, three-camera, and world model generated-view settings. As shown in Figure~\ref{fig:real_robot_tasks}, the default agent-view misses action-critical object cues in both tasks. In \emph{Oreo-to-Plate}, the robot grasps an Oreo and places it on a plate hidden behind a rice cooker. In \emph{Occluded-Doll Move}, the robot grasps a doll partially blocked by a box and moves it into another container.

\begin{wrapfigure}{r}{0.38\linewidth}
  \centering
  \vspace{-0.7em}
  \includegraphics[width=\linewidth]{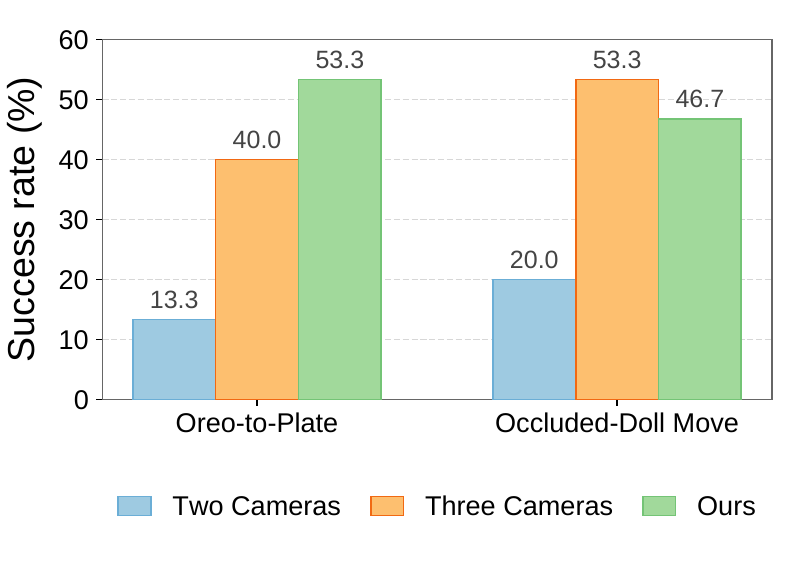}
\caption{\textbf{Real-robot performance over 15 trials per task.}}
  \label{fig:real_robot_view_config}
  \vspace{-2.0em}
\end{wrapfigure}

We compare three deployment configurations: two-camera, three-camera, and UniviewVLA. As shown in Figure~\ref{fig:real_robot_view_config}, the two-camera policy succeeds in only 13.3\% of Oreo-to-Plate trials and 20.0\% of Occluded-Doll Move trials. Adding a physical third camera improves success to 40.0\% and 53.3\%, respectively, confirming the importance of additional visual coverage in real-world occlusion tasks. UniviewVLA achieves 53.3\% and 46.7\% with only the two default cameras, substantially outperforming the two-camera policy and approaching the three-camera setting without the hardware cost of an extra calibrated camera. These results demonstrate that the world model's compact multiview token prediction and action-entropy view selection transfer to real-world deployment, providing a favorable return on investment (ROI) between occlusion robustness and deployment cost.

\section{Conclusion}

\label{sec:conclusion}

We presented UniviewVLA, a unified multiview VLA framework with world-modeling capabilities that predicts multiview future scene evolution for action prediction from standard agent-view and wrist-view observations, decoupling auxiliary-view reasoning from deployment-time camera requirements. To make world model generated views efficient for closed-loop robot manipulation, UniviewVLA compresses each generated view from 625 to 16 tokens and selects the most action-informative view via action entropy. Across simulation and real-robot evaluations, UniviewVLA maintains excellent occlusion-free manipulation performance while improving occlusion-task performance with only two standard deployed cameras. In addition, this work contributes six occlusion-focused tasks and a complete multiview data collection, processing, and evaluation pipeline for robot learning. These results demonstrate that UniviewVLA provides a practical framework for improving occluded robot manipulation without adding deployment-time cameras by leveraging world model capabilities to predict future scene dynamics.

\section{Limitations}
\label{sec:limitations}

UniviewVLA shares common constraints of generative robot policies. Auxiliary-view generation adds modest inference overhead, and performance depends on the coverage of multiview training data. Our evaluation focuses on tabletop and occlusion-focused tasks; extending to more diverse scenes, mobile viewpoints, and longer-horizon manipulation remains future work.
%===========================================================================

\bibliography{main}

\clearpage
\appendix

\section{Implementation details}
\label{sec:appendix_implementation}

\paragraph{Stage 1: multiview world-model post-training.}

All UniviewVLA experiments use models trained on 8 NVIDIA H-series GPUs with bf16 and DeepSpeed ZeRO-3.
We use the same Emu3-based autoregressive backbone as UniVLA, together with a VQ visual tokenizer for full-view prediction.
In Stage 1, the world model is trained to autoregressively predict next-frame VQ tokens of auxiliary workspace views from the agent view, wrist view, and language instruction, learning to predict multiview future scene evolution from standard observations. Each full view is represented as $25\times25$ VQ tokens. Training runs for 30K steps with global batch size 8 and a cosine learning rate schedule from $8\times10^{-5}$ to $5\times10^{-6}$.

\paragraph{Stage 2: action fine-tuning.}
Stage-2 fine-tuning uses a FAST action tokenizer for action prediction and runs for 24K steps on LIBERO and CALVIN and 30K steps on real-robot tasks. To prevent the model from collapsing different auxiliary views into the same action-conditioned supervision and to enable action-entropy view selection at inference, each demonstration is paired with a view-specific prompt prefix of the form ``predict \{view\} view and \{instruction\}''. The same prompt format is used for training and entropy-based inference-time view selection. For example, the instruction ``open the bottom drawer of the cabinet and put the bowl in it'' is rewritten as ``predict 60deg view and open the bottom drawer of the cabinet and put the bowl in it''.

\paragraph{Inference: entropy-based dynamic view selection.}
UniviewVLA uses the same view-specific prompt format as in Stage 2, selects among the candidate auxiliary views by minimizing action entropy, and re-selects every 30 policy steps, enabling adaptive view selection during execution.

\section{Multiview Tasks and Data Details}
\label{Tasks Details}

\begingroup
\setlength{\textfloatsep}{6pt plus 1pt minus 2pt}
\setlength{\floatsep}{6pt plus 1pt minus 2pt}
\setlength{\intextsep}{6pt plus 1pt minus 2pt}
\setlength{\abovecaptionskip}{2pt}
\setlength{\belowcaptionskip}{0pt}

We provide additional details for the multiview data collection and customized occlusion-focused tasks used in Section~\ref{sec:occlusion_eval}. For standard simulation benchmarks, we obtain multiview supervision by replaying recorded demonstrations with additional workspace cameras in the simulator. LIBERO uses $60^\circ$, $120^\circ$, $240^\circ$, $300^\circ$, and overhead views, as shown in Fig.~\ref{fig:multiview_row_libero}. Since the rear tabletop side in CALVIN is mostly redundant or weakly informative, we use more widely spaced views at $30^\circ$, $216^\circ$, $288^\circ$, and overhead, as shown in Fig.~\ref{fig:multiview_row_calvin}.

\begin{figure}[H]
\centering
\includegraphics[width=\linewidth]{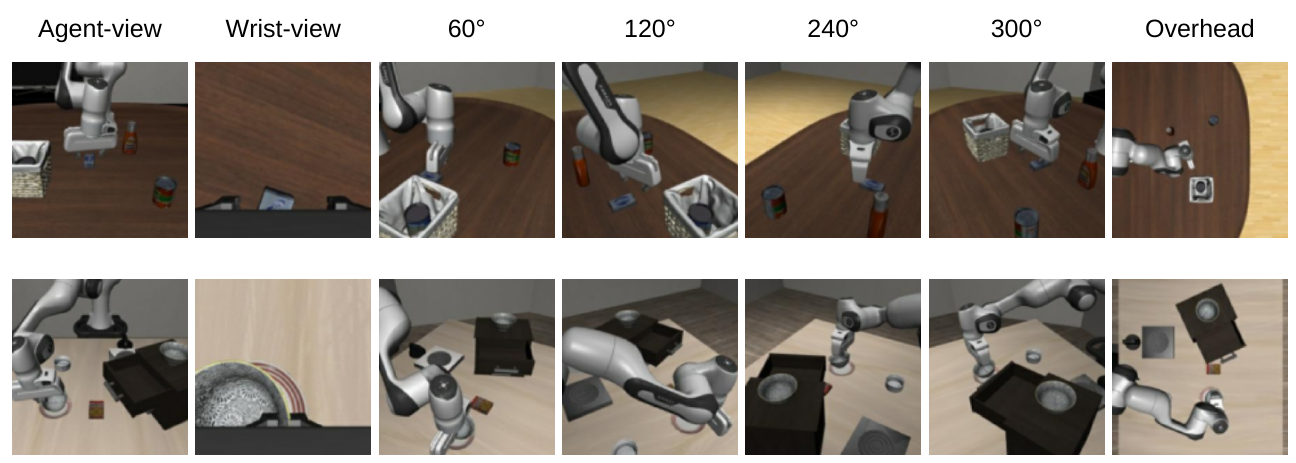}
\caption{
\textbf{LIBERO multiview.}
}
\label{fig:multiview_row_libero}
\end{figure}

To further evaluate the importance of multiview information, we construct six customized occlusion-focused tasks that hide action-critical cues from the standard viewpoints. Specifically, we follow the LIBERO BDDL format to design occluded manipulation scenes, as shown in Fig.~\ref{fig:6mask_task}, and collect multiview demonstrations with a hardware-in-the-loop teleoperation setup based on a 3D-SpaceMouse. Each task contains 60 demonstrations. Table~\ref{tab:occlusion_tasks5} lists the corresponding language instructions. Task 1 requires the robot to open a drawer that is partially hidden by the occluder and place the bowl inside it. Task 2 requires placing a bowl onto a plate occluded by the drawer. Task 3 requires grasping a wine bottle and placing it on a wine rack hidden by a box. Task 4 requires placing an occluded bowl into the drawer and closing it. Task 5 requires placing a mug onto a plate partially occluded by a microwave door. Task 6 requires turning off a switch occluded by the moka pot and then placing the moka pot on top of the cabinet.

\begin{figure}[H]
\centering
\includegraphics[width=0.92\linewidth]{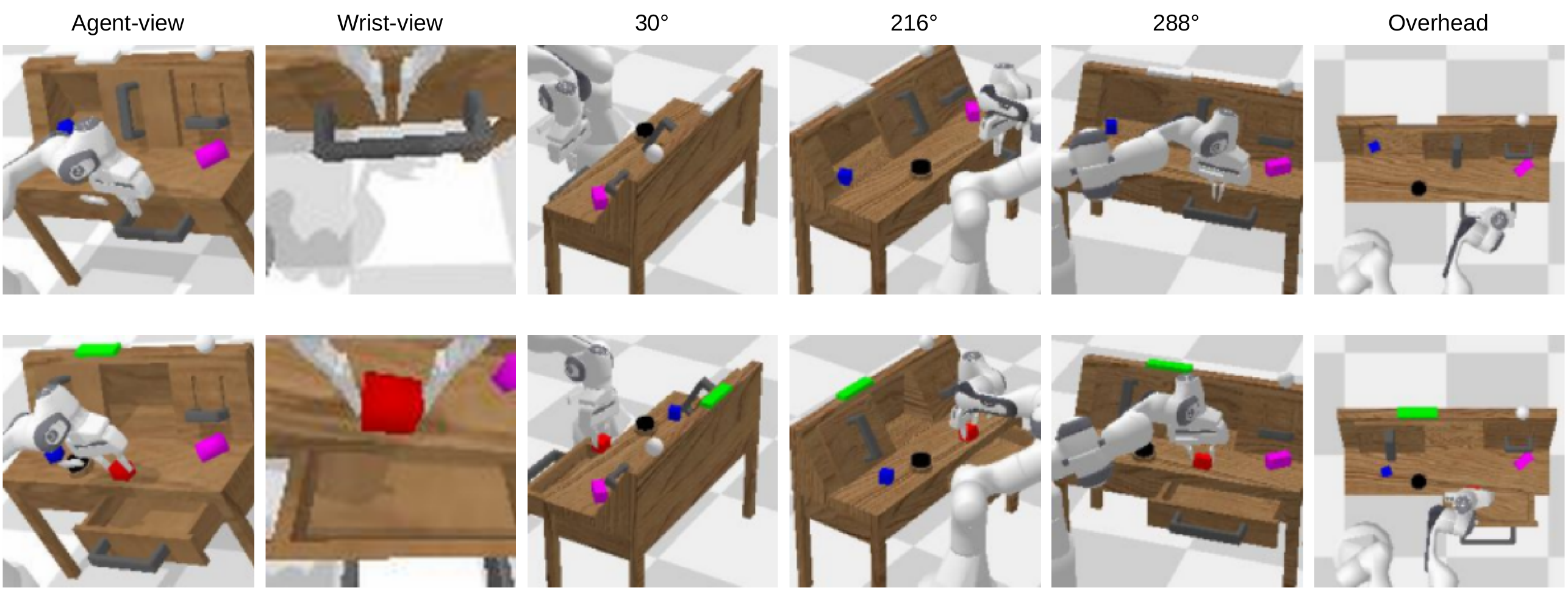}
\caption{
\textbf{CALVIN multiview.}
}
\label{fig:multiview_row_calvin}
\vspace{-6pt}
\end{figure}

\begin{table}[H]
  \centering
  \caption{%
    \textbf{Language instructions for six customized occlusion tasks.}
  }
  \label{tab:occlusion_tasks5}
  \small
  \setlength{\tabcolsep}{4pt}
  \begin{tabular}{@{}c l p{0.58\linewidth}@{}}
    \toprule
    ID & Task & Language Instruction \\
    \midrule
    1 & Scene1 Bowl & open the bottom drawer of the cabinet and put the bowl in it \\
    2 & Scene2 Bowl & put the black bowl on the left plate \\
    3 & Scene4 Wine & pick up the wine bottle at the back and put it on the wine rack \\
    4 & Scene4 Drawer & put the black bowl at the left in the bottom drawer of the cabinet and close it \\
    5 & Scene7 Mug & put the yellow and white mug on the plate \\
    6 & Scene8 Moka & turn off the stove and put the moka pot on top of the cabinet \\
    \bottomrule
  \end{tabular}
\vspace{-4pt}
\end{table}

\begin{figure}[H]
\centering
\includegraphics[width=0.92\linewidth]{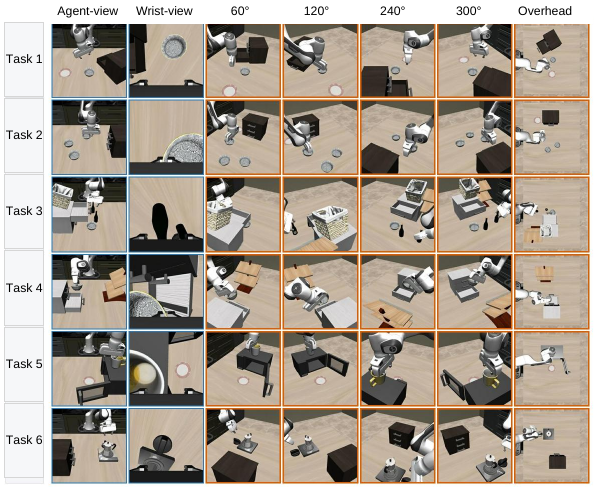}
\caption{%
\textbf{Six customized occlusion-focused tasks.}
Each task hides action-critical state cues from the default agent-view
camera while preserving the same two deployed physical observations.
}
\label{fig:6mask_task}
\vspace{-6pt}
\end{figure}

For real-robot evaluation, to avoid costly hardware and data-interface modifications, we design right-arm-only occlusion tasks on a Mobile ALOHA dual-arm platform. Accordingly, the unused left-wrist camera is repurposed as an additional side-view camera by placing it at a task-specific viewpoint using an extension cable. We collect 100 demonstrations for each of the Oreo-to-Plate and Occluded-Doll Move tasks. The real-robot multiview setup is shown in Fig.~\ref{fig:multiview_row_real}.

\begin{figure}[H]
\centering
\includegraphics[width=0.88\linewidth]{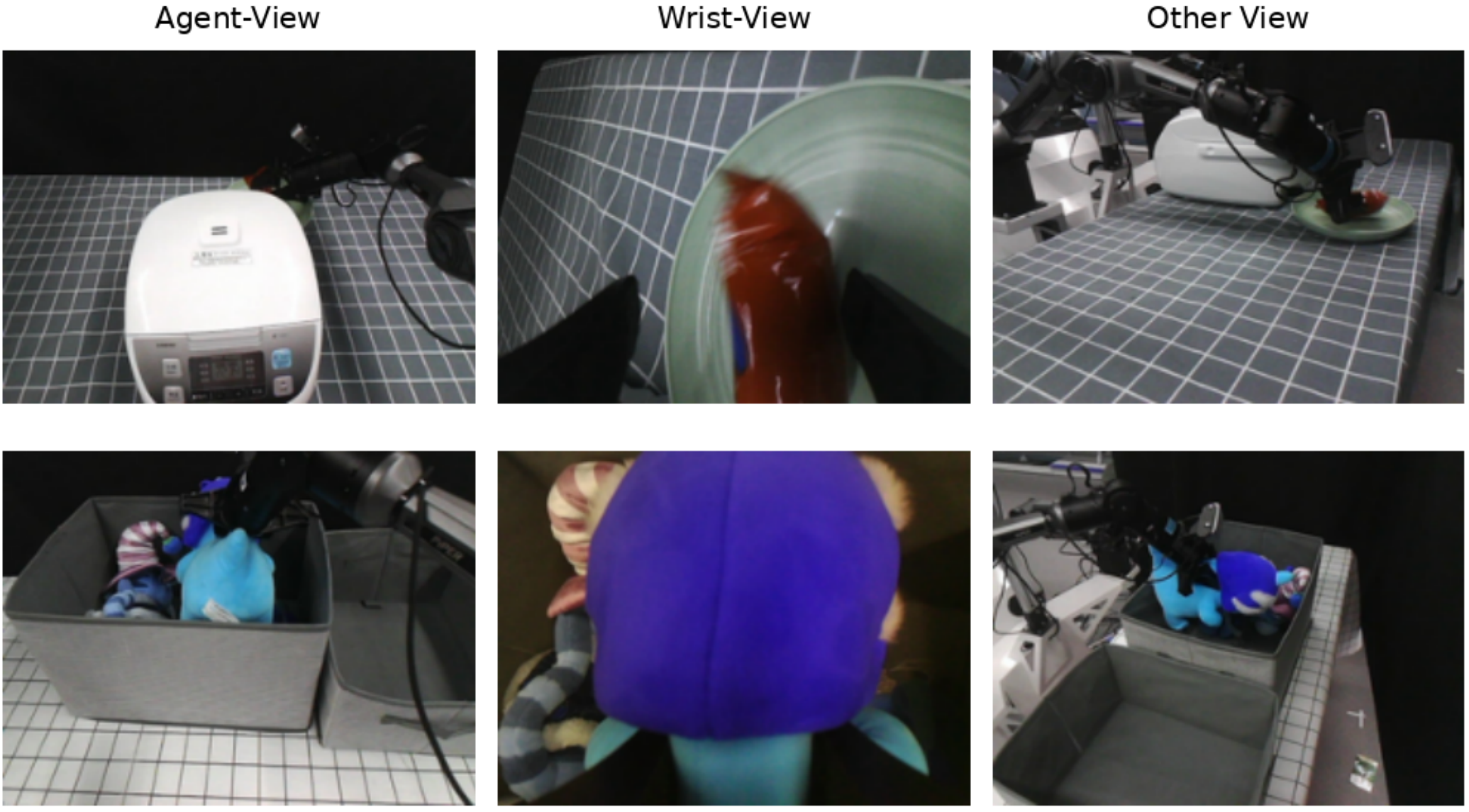}
\caption{%
\textbf{Real-robot multiview.}
}
\label{fig:multiview_row_real}
\end{figure}

\endgroup

\end{document}